# Automatic Detection and Classification of Waste Consumer Medications for Proper Management and Disposal

Bahram Marami* and Atabak Reza Royaee

DispoMeds, LLC, New York, NY, USA

**Abstract:**

Every year, millions of pounds of medicines remain unused in the U.S. and are subject to an "in-home disposal," i.e., kept in medicine cabinets, flushed in toilet or thrown in regular trash. In-home disposal, however, can negatively impact the environment and public health. The drug take-back programs ("drug take-backs") sponsored by the Drug Enforcement Administration (DEA) and its state and industry partners collect unused consumer medications and provide the best alternative to in-home disposal of medicines. However, the drug take-backs are expensive to operate and not widely available. In this paper, we show that artificial intelligence (AI) can be applied to drug take-backs to render them operationally more efficient. Since identification of any waste is crucial to a proper disposal, we showed that it is possible to accurately identify loose consumer medications solely based on the physical features and visual appearance. We have developed an automatic technique that uses deep neural networks and computer vision to identify and segregate solid medicines. We applied the technique to images of about one thousand loose pills and succeeded in correctly identifying the pills with an accuracy of 0.912 and top-5 accuracy of 0.984. We also showed that hazardous pills could be distinguished from non-hazardous pills within the dataset with an accuracy of 0.984. We believe that the power of artificial intelligence could be harnessed in products that would facilitate the operation of the drug take-backs more efficiently and help them become widely available throughout the country.

1. Introduction

    1.1 Motivation

Every year, about five billion prescriptions are filled by retail pharmacies in the U.S. [1]. Various studies show that not all prescribed medicines are consumed [2]. The overwhelming majority of unused consumer medications are either stored in medicine cabinets or disposed of improperly [2]. The unused consumer medications pose serious public health and environmental risks. According to U.S. Substance Abuse and Mental Health Services (SAMHSA), an estimated 18 million people have misused medications in 2017 [3]. The majority of those who misused prescription medications reported obtaining them from the medicine cabinets of friends and family [4]. There are over 100 prescription drug related deaths in the U.S. each day [5]. In a study, over ninety percent of the people surveyed indicated that they store their unused drugs at home, of which, 17.4% stored them indefinitely, and 18% flush and 62.7% discarded them in regular trash [2]. Flushing or trashing medicines, however, have resulted in pharmaceuticals ending up in waterways of majority of the states [6,7,8]. In general, the excess medications in the hand of consumers creates risks for unintentional poisoning, drug abuse or misuse and environmental contamination.

The best solution for preventing these risks is by practicing proper out-of-home disposal of unused consumer medications. Drug take-back programs ("drug take-backs") operated by the

---

*corresponding author, Email: bahram.marami@dispomeds.com

Drug Enforcement Administration (DEA) and its state, local and industry partners are slowly emerging in the U.S. as an alternative to in-home disposals of medicines. The drug take-backs include take-back events, take-back kiosks and drug mail-backs. They provide an environmentally friendly and safe options for consumers to dispose of their unused, expired or unwanted medications. For example, each year DEA conducts two one-day national take-back events at over 6,000 of its take-back sites across the country and collects about 500 tons (one million pounds) of unused medications each day [9]. Almost every police station in the country has a mail-box type drug take-back kiosk for collecting unused consumer medications around the clock. Chain pharmacies such as CVS, Walgreens and Rite Aid also provide drug take-back kiosks at their select locations. Furthermore, Med-Project [10] an affiliate of Pharmaceutical Research and Manufacturers of America (PhRMA) currently operates take-backs in a few states (CA, OR, WA, NY and MA). The DEA, Med-Project and chain pharmacies collect over 1000 tons (over two million pounds) of unused consumer medications at their take-back events every year [11]. However, there are two main shortcomings with the current drug take-backs. They are (a) low consumer participation and (b) high drug disposal costs.

Currently, only about 10% of the consumers participate in drug take-backs [2]. The consumers who participate in these programs are often those who are deeply concerned about the negative public health and environmental impacts of unused consumer medications and are willing to be inconvenienced if necessary but do the right thing and return their unused medications to drug take-backs for proper disposal. For the remaining 90% of the consumers, there appears to be a lack of knowledge about, access to, or motivation to participate in, the drug take-back programs. However, as the drug take-backs become more widespread, the consumer participation in these programs should increase.

The second drawback of the drug take-backs is their high cost of operation. The high operation cost stems from the fact that the medications collected by the drug take-backs are commingled. About 10% of the medications are either hazardous (e.g., Warfarin, Physostigmine, Nicotine, Arsenic Trioxide, Cepastat Lozenges, Chloraseptic Spray, Reserpine, Selenium Sulfide, etc.) by the EPA regulations [12] or regulated (opioid containing medications such as oxycontin) by the DEA regulations [13]. Since the drug take-backs currently do not have the capability to prevent the commingling of hazardous and DEA-regulated drugs and non-hazardous, non-DEA-regulated drugs, the entire volume of the medications collected at the take-backs is considered hazardous and DEA-regulated. This in turn excessively increases the cost of handling, storage, transportation, and disposal of the waste generated by the drug take-backs. For instance, the EPA recommends that the entire volume of the medications collected at the drug take-backs be disposed of at EPA-approved commercial hazardous waste incinerators [14], of which there are only eight in the entire U.S. and mostly located in Midwest states. These locations include Clean Harbors Environmental Services with four locations (in El Dorado, Arkansas; Deer Park, TX; Kimball, NE; and Aragonite, UT), Veolia Environmental Services with two locations (in Sauget, IL and Port Arthur, TX), Heritage Environmental Services with one location (in East Liverpool, OH) and Ross Environmental Services with one location (in Grafton, OH). However, the operation cost of the drug take-backs (including the disposal and the associated transportation costs) must reduce if the drug take-backs are to be become sustainable and widespread.

A technique for preventing the commingling of the hazardous and DEA-regulated medications and non-hazardous medications could in theory remedy both of the above-mentioned shortcomings. Such technique will in effect results in a 90% reduction in hazardous wastes generated at the drug take-backs, which will then significantly reduces the associated transportation and disposal costs. Such technique will allow for the 10% hazardous and DEA-regulated drugs to be disposed of at more expensive commercial hazardous waste incinerators while the 90% non-hazardous drugs may be disposed of at less expensive waste-to-energy incinerators. In addition to making the drug take-backs cost effective, a technique for identification and separation of hazardous and non-hazardous drugs can generate important data about what, when and where medications are wasted. This data is needed for improving the public health and the supply chain efficiency.

In this paper, we showed that it is possible to automatically and accurately identify and distinguish hazardous and non-hazardous medications. This technology can address the above-mentioned shortcomings by offering a new way of collecting and disposing of waste medications. It can help expand consumer drug take-back programs by reducing their operation costs and may allow for channelling some of the savings to consumers in form of incentives to encourage more consumer participations in the drug take-backs. The application of AI and computer vision to drug take-backs and preventing the hazardous and DEA-regulated medications from being commingled with non-hazardous medications promotes a more proper disposal of waste, fosters efficiency and environmental sustainability and contributes to a reduction in prescription drug abuse/misuse epidemic.

### 1.2. Previous work on identification of medications

Thanks to the computational power provided by modern CPUs and GPUs in the past ten years, deep learning-based methods have shown superb performance in solving large-scale computer vision problems. For automatic identification of prescription medications, several computer vision and machine learning based methods have been proposed in recent years to enhance patients' care and their safety. These automatic methods identify medications only based on their visual appearance; therefore, unlike platforms such as Drugs.com [15] and WebMD [16], they do not require the user to manually submit the necessary pill characteristics. To encourage the development of algorithms and software for automatic identification of consumer images of prescription pills and match to their corresponding reference images, the United States National Library of Medicine announced a challenge and provided 7,000 images from 1,000 pills [17]. Eleven different teams submitted their solutions, among which four teams utilized deep learning methods and seven teams used traditional machine learning and computer vision methods based on extracted texture, color, shape, etc feature vectors from the image content [18]. The winner of the competition [19] developed a deep learning system to recognize consumer-quality images. The model had a small footprint enough to be deployed into a mobile device. To localize the pill in the consumer images, they used the histogram of oriented gradients as features and trained a support vector machine (SVM) based detector that identifies the pill location. Then, they utilized a multi-CNNs (convolutional neural networks) architecture capturing color, shape, texture and imprint characteristics of the pills for classification. With their multi-CNNs model, they achieved a top-5 accuracy of 0.831 and 0.964 using one-side and two-sides of the pills, respectively [19].

Following the conclusion of the challenge, different techniques have been proposed for automatic pill recognition using the NLM or other pill image datasets. Wang *et al.* [20] used an edge detection method for localization and trained three different GoogLeNet Inception Networks [21] specialized on color, shape and features for classification using NLM dataset. Wong *et al.* [22] used an AlexNet [23] type architecture for the classification of images obtained from 400 commonly used tablets and capsules in a controlled environment using cameras of two mobile devices. In another study, Ou *et al.* [24] used a feature pyramid network [25] for simultaneous detection and classification of images obtained from 131 different pills. Most recently, Delgado *et al.* [26] proposed and validated a deep learning-based method for pill image recognition using NLM image dataset. They first used a fully convolutional network (FCN) [27] for detection and localization of the pill blobs through segmentation in the images. Then, they trained and validated different CNNs for classification of the localized pills in the images.

There are several commercial products for medication identification based on pill images in the market. The MedEye Pill-Scanning system by Mint Solutions [28] and identRx by PerceptiMed [29] are two similar scanning devices, designed to help nurses, patients and clinicians correctly identify medications. These systems can identify several pills inserted into an enclosed pod using images obtained under controlled conditions. The MedSnap system [30] uses computer vision through a smartphone application to identify medications for counterfeit surveillance, supply chain security and production monitoring services. Moreover, MedGlasses by Chang *et al.* [31, 32] is a wearable smart-glasses-based system for visually impaired chronic patients to automatically recognize prescription medications.

However, none of the above commercial products has been applied to drug take-backs and management of medications waste. The collection and disposal of unused consumer medications by the DEA or Med-Project are regulated by the title 21 of Code of Federal Regulation, section 1317 (21 CFR § 1317) and does not prohibit the application of the method described herein to the drug take-backs so long as the drug collected are in custody and control of law enforcement. The application of the technique discussed here to the drug take-backs will be in compliance with 21 CFR 1317 as all drug take-back events and kiosks that operate in police stations are already in custody and control of law enforcement.

## 2. Method

### 2.1 Pill image dataset

We developed and tested a pill detection and classification technique on pill image dataset provided by the national library of medicine (NLM) as its pill image recognition challenge [17]. This challenge was announced and concluded in 2016, but the dataset is available to the public. The dataset contains 2,000 reference images (top row in Figure 1) of 1,000 pills (one from the front and one from the back) and 5,000 consumer-quality images of the same 1,000 pills (bottom row in Figure 1). The consumer images were obtained using various digital cameras in different camera angles and under various lighting conditions. Most of the consumer pill images contain a single pill, however, 2 (out of 5,000) contain two pill images (front and back). Therefore, for each consumer-quality image, there exists two reference images, one from the front and another from the back of the pill.

All the pills provided by the NLM are labeled using an 11-digit number (e.g., 00172-6359-60), called the national drug code (NDC), as a universal product identifier. There are only 924 unique NDCs among 1,000 different pill images provided by the challenge. Some of the pills in the dataset were assigned different labels, although they have the same NDC, but different physical appearance (shape, color, imprint). Thus, Delgado *et al.* [26] grouped the images with the same NDC together and classified the provided images into 924 classes. Since, the challenge is concluded, in order to validate the performance of their algorithm, Delgado *et al.* held out 20% of the consumer-quality images (1,000 out of 5,000 images) for testing and used the remaining 80% for training through cross-validation. In this paper, we compare the performance of our developed method to that of [26], using exactly the same list of test images that the authors of [26] provided us upon our request.

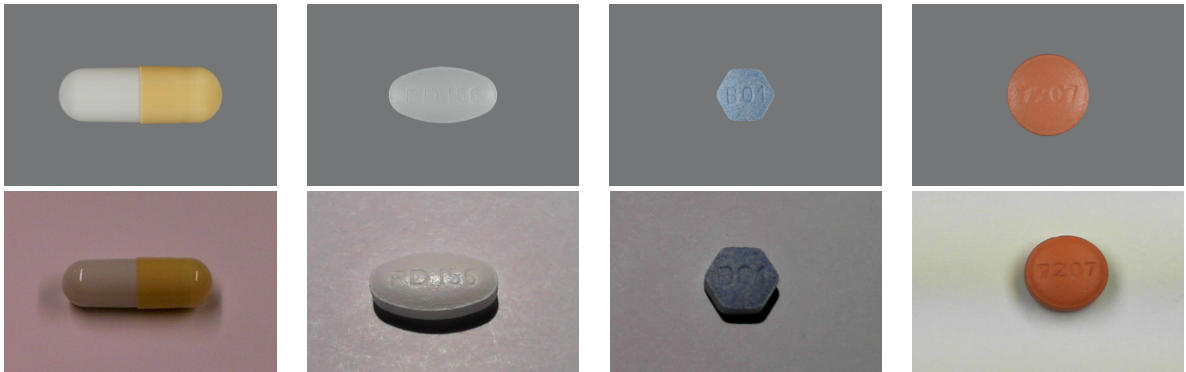

Figure 1. Reference (top row) and their corresponding consumer-quality (bottom row) medicine images

### 2.2 Pill detection through segmentation

The NLM pill dataset does not provide pill localization information or segmentation masks. To train our pill detection network, we generated 160,000 synthetic images using NLM's reference images. The background of the reference images has a fixed RGB value, e.g., 118,118,118. Using this fact, a foreground mask was created by removing the background, followed by binary opening and closing to remove noise. We collected over 1,000 publicly available background images, including desks, papers, fabrics, tables, carpets, as well as metal, wood and marble textures. Then, we randomly placed one or multiple extracted reference pill images and created 400×400 images. As shown in Figure 2, the pill images were placed in different orientations. After placement of the pills, a random 0-25 pixel drop shadow was added to the images. Moreover, a random perspective transformation and gaussian blurring were applied on the image.

We used a DeepLabv3+ with a Xception backbone [33] architecture to train a pill detector through segmentation, see Figure 3. This network achieved the highest segmentation accuracy on the PASCAL VOC dataset [34], although, other deep learning-based image segmentation models might have produced similar results for our application. To train the pill detector network, an Adam optimizer [35] with a learning rate of 0.001 was used. We used 120,000 of the generated synthetic images for training the pill detector. While training various image augmentation methods were applied, including rotation, flip, smoothing, sharpening, jpeg compression, perspective transformation as well as contrast and brightness adjustments.

During the inference on the consumer pill images, in order to localize the pills on the segmented image, all objects smaller than 100 pixels were removed, followed by a binary opening and closing to remove noise, and finally dilated with a kernel size of 15 pixels to account for cutoffs and drop shadows. Finally, all connected components (pills) were detected in the mask and localized with their bounding box.

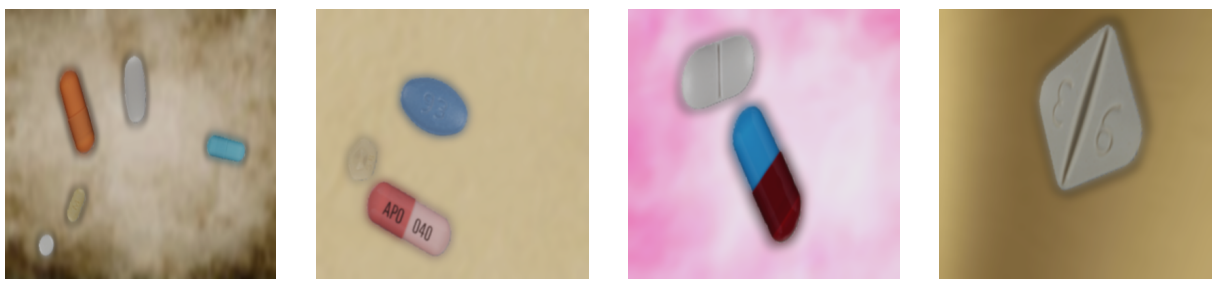

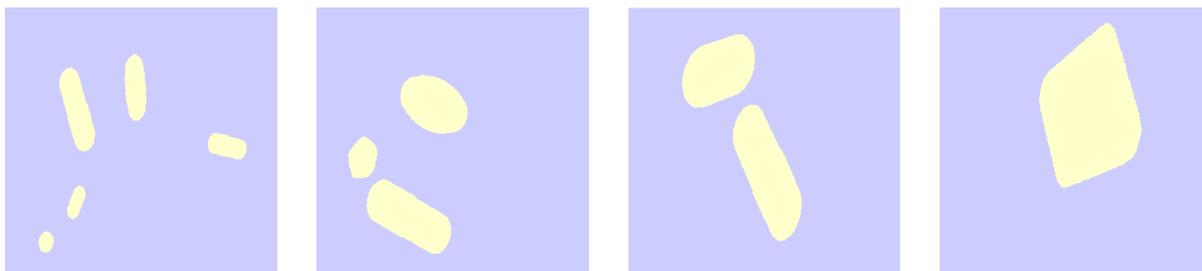

Figure 2. Examples of synthetic pill images produced from the reference pill images (top row) and their ground truth segmentation (bottom row).

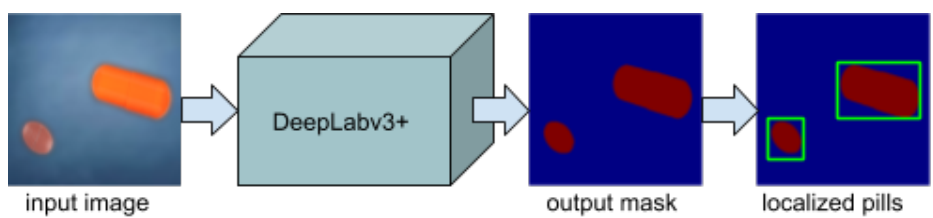

Figure 3. Detection and localization of pills in the images through image segmentation

### 2.3 Pill classification

We used the InceptionV4 [36] architecture to train a CNN to classify the detected medicines in the images into 924 classes. Using the localized pill images from the previous step, a 299×299 image is created for each pill present in the image, in which the pill is placed in the center, see Figure 4. The 299×299 pill images created from each consumer-quality image are then used to predict the label by the classifier. Since we had set aside 20% of consumer-quality images for testing, we only had 4,000 images (or about 4 images per class of medicines) to train our AI algorithm with. To avoid overfitting the training data, we further generated 80,000 synthetic images (in a similar fashion as described above, placing pills on background images, adding drop shadow, smoothing and perspective transformation) from the consumer-quality images and 80,000 synthetic images from the reference images. We also generated multiple scaled images from the consumer-quality images and the reference images totalling 220,010 medicine images that were used for training the CNN. As shown in Figure 4, we used a five-channel image (RGB+grayscale+gradient) as input. Also, we added four fully connected layers with dropouts and ReLU activation function at the end of the convolutional layers. The dropout was tuned to 30% through validation to avoid overfitting of the CNN to the training dataset. We trained the CNN within four cross-validation folds, exactly identical to those in [26], as well as 10 cross-validation in which the folds were randomly selected from the training dataset. Only consumer-quality images were divided for cross-validation and all synthetic images created from reference images were used in every fold for training. We trained the classifier using PyTorch with Adam optimizer [35], a learning rate of 0.00001, and a cross-entropy loss function.

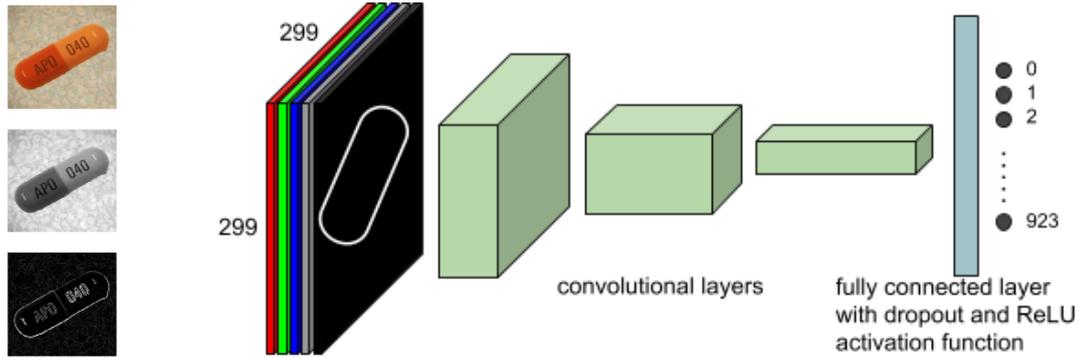

Figure 4. Convolutional neural network for the classification of customer-quality pills

## 3. Experimental Results

### 3.1 Pill detection through segmentation

We applied the trained pill detector network on the 20,000 synthetic pill images that were set aside for testing the segmentation network. The average Dice similarity metric between pill detector network's prediction and the ground truth mask was 0.989. After post-processing the predicted masks and removing noise and small connected components, the average Dice similarity metric was 0.993.

Using the Pill detector network, we also detected the pills on the 5,000 consumer-quality images. The detection algorithm was able to detect, localize and estimate the bounding box on 100% of the existent pills on the consumer-quality images, i.e. 5,002 pills on 5,000 images (see Figure 5 for examples). Since we didn't have the ground truth masks for these images, computing the Dice similarity metric was not possible.

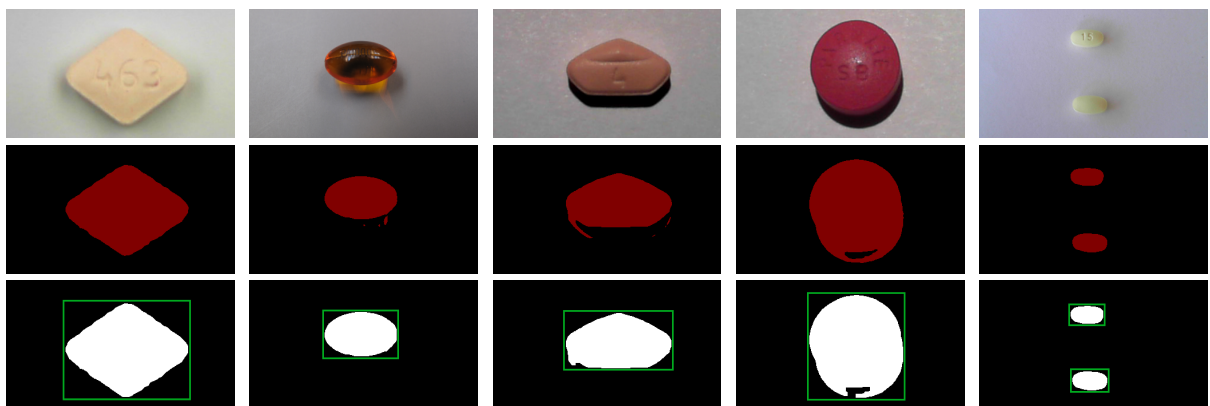

Figure 5. Examples of consumer-quality images (top row) and their segmented masks from the DeepLabv3+ (middle row) and the localized pill after post-processing with bounding boxes (bottom row).

### 3.2 Pill classification

We evaluated the performance of the trained classification networks on 1,000 consumer-quality test images. As shown in Table 1, the 10-folds ensemble of networks produced the best top-1 classification accuracy (0.912) and the 4-folds ensemble of networks resulted in the best top-5 classification accuracy (0.985). The accuracy of the best single classification network was 0.891 and the worst single classification network was 0.85. Moreover, the micro-average precision score on the test images was 0.9594 and 0.9658 using the 4-folds and 10-folds ensemble networks, respectively. The precision-recall curve plots of both 4-folds and 10-folds ensemble networks are given in Figure 6. This plot also shows that the average precision score of the 10-folds ensemble network is slightly better than that of the 4-folds ensemble network. The Venn diagram between the 4-folds and 10-folds ensemble networks prediction of 1,000 test images is also given in Figure 7. Examples of inaccurate predictions by either of both ensemble networks are given in Figure 8.

Table 1: Classification accuracy and the average precision score (micro) on the test images (1,000 consumer-quality test images)

|  |  | 4-folds | | | 10-folds | | |
| --- | --- | --- | --- | --- | --- | --- | --- |
|  |  | front (533) | back (467) | all | front (533) | back (467) | all |
| top-1 accuracy | min | 0.906 | 0.786 | 0.85 | 0.916 | 0.816 | 0.872 |
|  | mean | 0.920 | 0.809 | 0.868 | 0.923 | 0.831 | 0.88 |
|  | max | 0.934 | 0.827 | 0.875 | 0.931 | 0.846 | 0.891 |
|  | ensemble | **0.944** | **0.842** | **0.896** | **0.949** | **0.869** | **0.912** |
| top-5 accuracy | min | 0.983 | 0.940 | 0.963 | 0.983 | 0.949 | 0.967 |
|  | mean | 0.988 | 0.953 | 0.972 | 0.990 | 0.958 | 0.975 |
|  | max | 0.994 | 0.961 | 0.979 | 0.994 | 0.964 | 0.98 |
|  | ensemble | **0.998** | **0.970** | **0.985** | **0.996** | **0.970** | **0.984** |
| micro-average precision | ensemble | **0.9839** | **0.9237** | **0.9594** | **0.9852** | **0.9378** | **0.9658** |

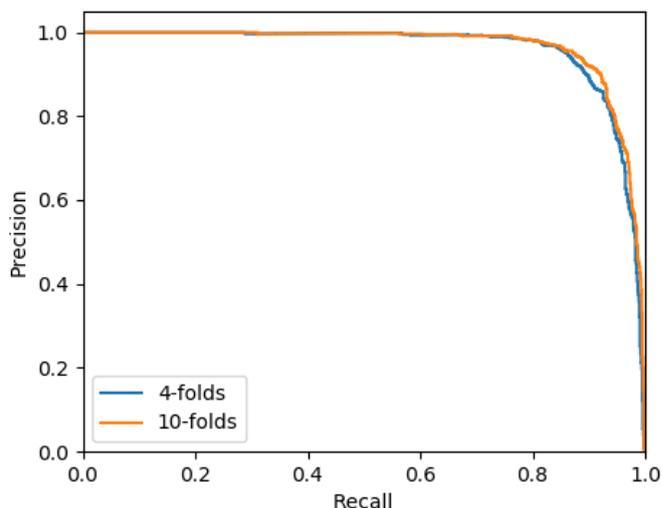

Figure 6. Precision-recall curve plots on the test set computed from micro-averages of the precision scores for each pill class resulted from 4-folds and 10-folds ensemble networks.

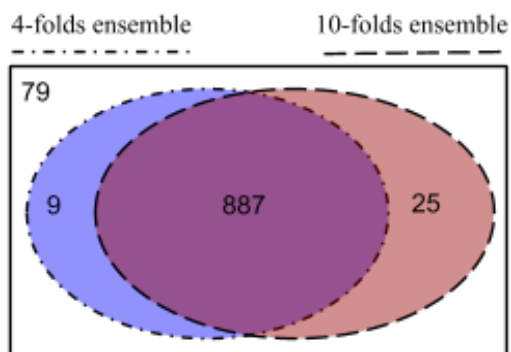

Figure 7. The Venn diagram between the 4-folds and 10-folds ensemble networks prediction of 1,000 test images. The diagram shows that 887 pill images were predicted accurately by both ensemble networks and 79 pill images were predicted inaccurately by both of them. 4-folds and 10-folds ensemble networks had different predictions on 34 pill images.

We reviewed all 1,000 test images, compared them to the reference pill images according to their labels and determined whether they contain pills with their front shown in the image or their back. 467 out of 1,000 images contained back of the pills, 532 contained front of pills and only one contained both front and back. For simplicity in representing the results, we considered this image in the group of the images containing pills with their front shown, however, its class was determined based on the average classification probability of them both. As shown in Table 1, the classification accuracy was on average about 0.1 higher in the images containing pills with their front shown than those containing pills with their back shown.

We compared the performance results of our developed algorithm to those of Delgado *et al.* [26] using an identical list of consumer-quality test (1,000 hold-out) images. Delgado *et al.* used four different CNN architecture to train four classifiers through a 4-folds cross-validation scheme. They chose the single best performing CNN on the cross-validation images for each architecture and used it to evaluate its performance on the test images. These networks and their performance are summarized in Table 2. A direct comparison of our best performing model (4-folds or 10-folds ensemble network) to those of [26] would not be fair. In Table 2, we only compare the results obtained from the single worst and the single best performing CNN in our 4-folds cross-validation training. It should be noted that we also used identical folds for cross-validation training to those of Delgado *et al.* [26]. As given in Table 2, the top-1 and top-5 accuracy as well as the average precision of our trained classifiers (both the worst and the best classifiers) were higher than those in [26].

Table 2. Classification accuracy and average precision score (micro) on the test images by Delgado *et al.* (from [26]) and a direct comparison to the single worst and the single best performing classifiers in our 4-folds cross-validation training. Number of parameters for each CNN architecture is also given.

|  | Delgado *et al.* | | | | Our method: InceptionV4 | |
| --- | --- | --- | --- | --- | --- | --- |
|  | ResNet50 | InceptionV3 | MobileNet | SqueezeNet | The Worst | The Best |
| top-1 accuracy | 0.770 | 0.763 | 0.771 | 0.562 | **0.850** | **0.875** |
| top-5 accuracy | 0.953 | 0.948 | 0.944 | 0.832 | **0.963** | **0.979** |
| average precision | 0.8587 | 0.8594 | 0.8429 | 0.6133 | **0.9179** | **0.9372** |
| # of parameters (millions) | 26.35 | 26.56 | 5.04 | 2.06 | 42.6 | |

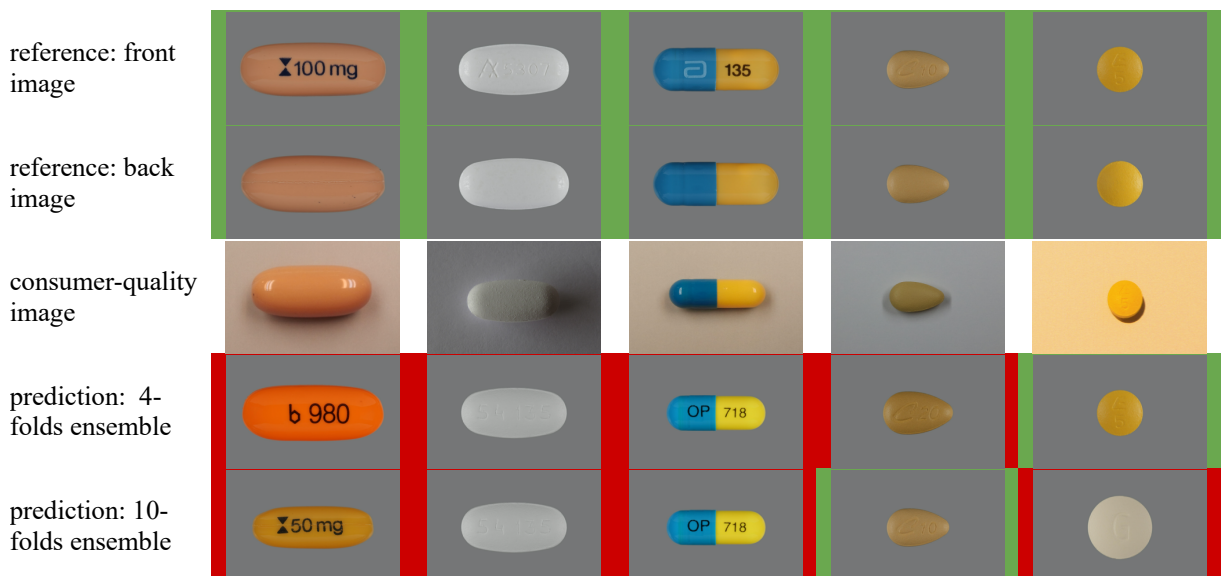

Figure 8. Examples of inaccurate prediction by 10-folds ensemble networks (first column from right), 4-folds ensemble networks (second column from right) or both ensemble networks (third, fourth, fifth column from right). The green background represents an accurate prediction (matching with the reference images) and the red background represents an inaccurate prediction.

As shown in Figure 8, in the majority of the misclassified pills, the consumer-quality image was taken from the side of the pill with no imprint or engravement. According to the file names of the reference images, this side is usually called "back" of the pill when the imprint only exists on one side of the pill. Hence, we evaluated the effect of having both the front and the back of the pill on the image in another experiment. To this end, we generated 2,000 synthetic images using the reference images, 1,000 from the front and 1,000 from the back of the pills. We extracted the largest empty background rectangle area from the 1,000 consumer-quality test images. We then used these images as background and placed the front and back of one of the reference pills images at each of them. The reason for using the background images of the consumer-quality test images was to make sure that the background images of the generated synthetic images for this experiment are totally different from those public images we downloaded and used in the training process. As explained in the previous section, we added random rotations, perspective transformations, blurring and contrast variations. Then, we applied the classification algorithms on these images to predict their labels. The front and the back images of a few example pills are given in Figure 9. The performance of the 4-folds and 10-folds ensemble networks in classifying these images is compared in Table 3. The classification accuracy was higher using only the images from the front of the pills than using only the images from the back of the pills. Also, using the average probability of both the front and the back pill images (front+back) improved the classification accuracy, significantly.

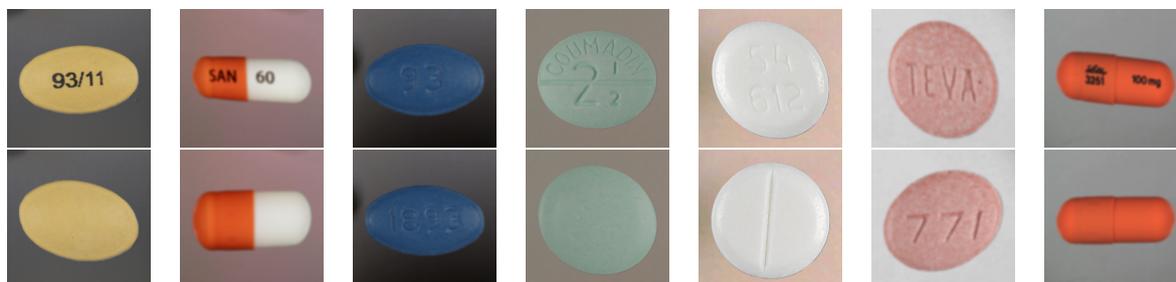

Figure 9. The front (top row) and the back (bottom row) of the pill created synthetically from the reference pill images. The size of these images is 299×299 and they are directly used for classification, without applying the detection network.

Table 3. Classification accuracy and average precision score (micro) on 1,000 synthetic images showing both the front and the back of the pill; front+back means the average probability of front and back images.

|  | 4-folds ensemble | | | 10-folds ensemble | | |
| --- | --- | --- | --- | --- | --- | --- |
|  | front | back | front+back | front | back | front+back |
| top-1 accuracy | 0.973 | 0.954 | 0.986 | 0.963 | 0.947 | 0.986 |
| top-5 accuracy | 0.994 | 0.991 | 0.998 | 0.993 | 0.986 | 0.996 |
| average precision | 0.9924 | 0.9872 | 0.9968 | 0.9892 | 0.9796 | 0.9946 |

We also analyzed the performance of the classifiers in separating the pills based on their disposal requirements under DEA and/or EPA regulations into two classes, hazardous and non-hazardous. We reviewed all 1,000 medicines and found out that 177 of them are hazardous and the remaining 823 are non-hazardous. The performance of both 4-folds and 10-folds ensemble networks in classifying the test images into regulated and unregulated pills are summarized and given in Table 4 and Table 5.

Table 4. Accuracy, recall, precision, f1-score and AUC (area under the receiver operating characteristics curve) of separating 1,000 test images into hazardous (177) and non-hazardous (823) drugs.

|  | 4-folds ensemble | | | 10-folds ensemble | | |
| --- | --- | --- | --- | --- | --- | --- |
|  | front | back | all | front | back | all |
| accuracy | 0.985 | 0.970 | **0.978** | 0.991 | 0.976 | **0.984** |
| recall (sensitivity) | 0.944 | 0.966 | **0.955** | 0.978 | 0.989 | **0.983** |
| precision | 0.965 | 0.885 | **0.923** | 0.967 | 0.897 | **0.930** |
| f1-score | 0.955 | 0.924 | **0.939** | 0.972 | 0.941 | **0.956** |
| AUC | 0.969 | 0.968 | **0.969** | 0.985 | 0.981 | **0.984** |

Table 5. Truth table for the prediction of 1,000 test images into non-hazardous and unregulated drugs, with the number true positives (TP), true negatives (TN), false positives (FP) and false negatives (FN)

|  |  | 4-folds ensemble | | 10-folds ensemble | |  |
| --- | --- | --- | --- | --- | --- | --- |
|  |  | Prediction | | Prediction | |  |
|  |  | non-hazardous | hazardous | non-hazardous | hazardous |  |
| Actual | non-hazardous | TN=809 | FP=14 | TN=810 | FP=13 | 823 |
|  | hazardous | FN=8 | TP=169 | FN=3 | TP=174 | 177 |
|  |  | 817 | 183 | 813 | 187 |  |

## 4. Discussions

The results presented in the previous section showed that advanced AI solutions can be applied for automatic identification and separation of prescription medications with a high accuracy, even with a limited available imaging dataset. The primary goal of this study was to achieve the highest possible classification accuracy using only four pill images per each pill type and show that hazardous medications can be separated from non-hazardous medications using the developed AI algorithm. Although we used a similar approach to that of Delgado *et al.* [26], as given in Table 2, our method outperformed theirs. The worst trained network in our method performed better than the best network in [26] on exactly the same test images. Because of the limited training data, overfitting was inevitable in this project. To mitigate the overfitting problem, we created 160,000 synthetic images from both consumer-quality and reference pill images with thousands of different background images in multiple scales. Also, it has been shown in multiple studies that the new InceptionV4 network (the architecture we used in this paper), which combines both Inception and ResNet blocks, performs better than any of the architectures used in [26]. We also added the grayscale and the gradient images to the RGB image and trained the networks with a 5-channel input image. However, our experiments showed this improved the classification performance only slightly over the 3-channel RGB input image.

The reason we followed Delgado *et al.* [26] in performing a 4-folds cross-validation training was to make sure that we have at least one pill type in the validation set for each fold, i.e., three pills in the training and one pill in the validation for each fold. Therefore, in the 10-folds cross-validation training, the validation set at each fold did not include all pill types. Nonetheless, the 10-folds cross-validation training, using 20% more training images, performed better than 4-folds (individual and ensemble networks). This also shows how the classification networks were prone to overfitting when there was a limited number of available images for training.

We analyzed all 113 test images that were predicted inaccurately by either of the ensemble networks. Other than being blurry, partially bright or dark, or distorted and having large shadows in the consumer-quality images, the main reason for misclassification in most cases is that the shape, size and the color of the pill is very similar to those of another pill type. As shown in Figure 8, the misclassified cases have very close (identical in some cases) shape, size and color, and the only differentiating feature is the imprint on the front or back of the pill. However, in the majority of the misclassified test images, the consumer-quality image had no imprint, or the imprint faded due to the image artifacts. In order to correctly identify these pills by patientcare or safety applications, high-quality images from both sides of the pills have to be provided by the users. To avoid errors in such situations, one solution could be grouping very similar pills (in terms of color, shape and size) into one class; then, separating the group into the individual types in another stage in a more controlled condition and using images from different sides of the pills.

Using a workstation with a Nvidia GeForce GTX 1080 GPU and Intel® Xeon(R) CPU E5-2620 v4 @ 2.10GHz CPU, the average computation time for pill detection through segmentation and post processing for localization was 0.11 sec, and the average classification time was 0.09, 0.20, 0.42 sec for a single network, 4-folds ensemble and 10-folds ensemble inference, respectively.

In this paper, we showed that ensemble of multiple networks as well as using both front and back images of the pills improves the classification accuracy. To build a product prototype for use at drug take-back events, we plan to investigate the use of multiple cameras to be able to obtain images from different directions and perspectives and develop an algorithm that uses multiple input images for identifying the pills. Furthermore, in order to reduce the inference time, we will use techniques such as single shot multibox detector [37], Faster R-CNN [38] or fully convolutional one-stage (FCOS) object detection [39] for simultaneous detection and classification. These networks are also small enough to be deployed into devices such as Nvidia Jetson AGX Xavier [40] module in order to have an AI-powered autonomous machine for separating medications at the point of collection.

## 5. Conclusions

This study provided a deep learning-based method for automatic detection and classification of prescription medications based on their images obtained by consumers using their cell-phone cameras. Our results showed that the method was capable of identifying the correct pills with high accuracy and average precision. It was also shown that the hazardous medications could be separated from non-hazardous medications with a high accuracy and recall (sensitivity). The application of the method described in this paper to the drug take-backs allows for prevention of hazardous and DEA-regulated medications from being commingled with non-hazardous medications. This in turn reduces the volume of hazardous waste generated by these programs and the associated transportation and disposal costs. In addition, since the described methods identifies the medicines, it is possible to collect accurate information about what medications are wasted in which geographical location within the country. Such information is valuable for promoting pharmaceutical supply chain efficiency as well as other applications such as prescription drug monitoring and drug diversion control.

## Acknowledgments

The authors would like to sincerely thank Dr. Hugo Fernandes and Dr. Nishan S. Mann for reviewing the manuscript and providing constructive comments and suggestions that improved the paper significantly.